\documentclass[10pt,leqno]{amsart}
\usepackage{graphicx}
\pdfoutput=1
\usepackage{indentfirst,csquotes}

\usepackage{amsmath,amssymb,amsfonts}
\usepackage{algorithmic}
\usepackage{graphicx}
\usepackage{textcomp}
\usepackage{xcolor}
\def\BibTeX{{\rm B\kern-.05em{\sc i\kern-.025em b}\kern-.08em
    T\kern-.1667em\lower.7ex\hbox{E}\kern-.125emX}}

\usepackage{algorithm}
    
\begin{document}

\title[Flexible K Nearest Neighbors Classifier]{Flexible K Nearest Neighbors Classifier:\\Derivation and Application for Ion-mobility\\Spectrometry-based Indoor Localization}
\author[Müller]{Philipp Müller*}

\let\thefootnote\relax
\footnotetext{*Faculty of Information Technology and Communication Sciences, Tampere University, Tampere, Finland, Email: philipp.muller@tuni.fi

Version: 13 March 2024

\copyright 2023 IEEE. Personal use of this material is permitted. Permission from IEEE must be obtained for all other uses, in any current or future media, including reprinting/republishing this material for advertising or promotional purposes, creating new collective works, for resale or redistribution to servers or lists, or reuse of any copyrighted component of this work in other works.

DOI: https://doi.org/10.1109/IPIN57070.2023.10332541

This work was supported in part by the Academy of Finland, grant 323472 (under consortium “GaitMaven: Machine learning for gait analysis and performance prediction”).}

\maketitle

\begin{abstract}
The K Nearest Neighbors (KNN) classifier is widely used in many fields such as fingerprint-based localization or medicine. It determines the class membership of unlabelled sample based on the class memberships of the K labelled samples, the so-called nearest neighbors, that are closest to the unlabelled sample. The choice of K has been the topic of various studies and proposed KNN-variants. Yet no variant has been proven to outperform all other variants. In this paper a new KNN-variant is proposed which ensures that the K nearest neighbors are indeed close to the unlabelled sample and finds K along the way. The proposed algorithm is tested and compared to the standard KNN in theoretical scenarios and for indoor localization based on ion-mobility spectrometry fingerprints. It achieves a higher classification accuracy than the KNN in the tests, while requiring having the same computational demand.
\end{abstract}


\section{Important note by the author}

As of October 2023 the author has been aware that \textbf{the concept behind the introduced Flexible K Nearest Neighbors classifier has been proposed and used before under the name Fixed Radius Near Neighbor search or Radius Neighbor classifier}. According to a survey by Bentley~\cite{Bentley1975}, the method was first used in 1966 by Levinthal in an "interactive computer graphics study of protein molecules", but from Levinthal's paper~\cite{Levinthal1966} neither a clear mentioning nor a derivation of the Fixed Radius Near Neighbor technique could be found. Hence it is unclear to the author where the underlying idea of the Flexible K Nearest Neighbors classifier was proposed for the first time.

To the authors knowledge, \textbf{the application of the concept for ion-mobility spectrometry fingerprint-based indoor localization is novel}. In addition, the \textbf{paper provides an insight into the strengths and the weaknesses of using only training samples within a predefined distance to the test sample for inferring the latter one's label}, which might be of interest for readers new to fixed radius-based nearest neighbors classifiers. Recent articles using the concept include, for example,~\cite{Krol2020},~\cite{Wang2020}, and~\cite{Grochowalski2021}.


\section{Introduction}

The $K$ Nearest Neighbors ($K$NN) classifier is a widely used and thoroughly studied machine learning algorithm due to its simplicity, ease of implementation and being parameter-free~\cite{Onyezewe2021}. It is an extension of the nearest neigbor rule, which is a suboptimal classifier whose error rate is at most twice the Bayes rate~\cite[p.\ 177]{Duda2001}.

In classification the label of so-called test samples is derived from the labels of so-called training samples. If the training and test samples follow the same distribution then support vector machines, Fuzzy and Bayesian classifiers, as well as neural-network-based classifier perform well. However, if the distributions of training and test samples vary even slightly then the performance of these methods can degrade significantly and $K$NN and its variants are commonly used~\cite{Bian2022}. The main idea of the $K$NN is to find the $K$ training samples that are closest, with respect to a predefined distance measure, to the test sample and infer the class membership from class memberships of these $K$ training samples.

One application where differences between distributions of training and test samples can be observed is fingerprint-based indoor localization. For example, in~\cite{Muller2014} the weighted $K$NN outperformed several parametric classifiers for indoor localization based on Wireless Local Area Network (WLAN) measurements if the WLAN access point density was high. However, if the density was low then the accuracy level of the weighted $K$NN degraded considerably. One reason for the lowered performance was the lack of training sample locations in which the same access points were observed as in the test sample (see~\cite{Muller2014} for details). In such a case choosing a different $K$ might help to improve the accuracy and numerous $K$NN-variants have been proposed that focus on optimizing $K$.

However, in some situations even optimizing $K$ is of little use. If there are no training samples within the close neighborhood of the test sample than any $K$NN-variant will yield a wrong class membership label for the test sample. Examples of such situations can be found, for example, in indoor localization (test sample from a room or area of a building for which no training data has been collected) or medicine (classification of a disease for which no training data is available). In order, to handle also such situations well and avoid time-consuming optimizations this paper proposes a $K$NN-variant in which the allowed distance between training and test samples is limited from above. Only training samples that are within the limit are used for inferring the label of the test sample. This way, a different $K$ will be found for each test sample, and therefore the proposed classifier is named Flexible $K$NN (Flex$K$NN). If $K=0$ then the Flex$K$NN will simply yield information that no training sample is close enough to the test sample to provide a reliable estimate of class membership for the test samples. This is, in the authors opinion, more useful than an untrustworthy label provided by $K$NN and its existing variants. To the authors knowledge no such $K$NN-variant has been published in earlier papers.

The remainder of this paper is organized as follow. Section~\ref{sec:RelWorks} discusses the standard $K$NN in detail and provides an overview on proposed $K$NN-variants. The Flexible $K$NN is introduced and explained in detail in Section~\ref{sec:FlexKNN}. In Section~\ref{sec:Appl} the performances of Flex$K$NN and standard  $K$NN are compared for indoor localization based on ion-mobility spectrometry (IMS) measurements. Section~\ref{sec:Concl} contains concluding remarks and an outlook on further research.


\section{Related work}\label{sec:RelWorks}

The standard $K$ Nearest Neighbors classifier is a model-free classifier that returns a label for an $n$-dimensional vector $\mathbf{x} = [ x_{1}, \ldots , x_{n} ] \in \mathbb{R}^n$, which indicates that the so-called test sample $\mathbf{x}$ belongs to a class $m$ ($m = \{1, \ldots M\}$). The label is derived by finding the $K$ closest samples from a database $\mathbf{Y}$ containing $N$ training samples $\mathbf{Y} = [\mathbf{y}_1, \ldots , \mathbf{y}_N] \in \mathbb{R}^{n \times N}$ for which their class affiliations (i.e. their labels) are known. The label that occurs most frequently amongst the labels of the K closest training samples (i.e. the K nearest neighbors) is then chosen as label for the test sample.

Closeness between test sample $\mathbf{x}$ and a training sample $\mathbf{y_i}$ is often measured by the Euclidean distance $d_{\text{E}}$, which is defined as
\begin{equation}
d_{\text{E}}( \mathbf{x}, \mathbf{y}_i ) = \sqrt{ \sum_{j=1}^{n} \left( x_{j} - y_{i,j} \right)^2}.
\label{eq:EuclideanDistance}
\end{equation}
For $\mathbf{x}, \mathbf{Y} \in \mathbb{R}^n$ Euclidean distance is a reasonable choice because $\mathbb{R}^n$ is a Euclidean space. For non-Euclidean spaces a different distance measure should be used. Furthermore, alternative distance measures have been proposed and tested also for use in Euclidean space. For example, \cite{Alkasassbeh2015} proposed the Hassanat distance, which uses maximum and minimum vector points. It outperformed the standard $K$NN and eight $K$NN-variants over eight machine learning benchmark datasets from the field of medicine in~\cite{Uddin2022}. One of the $K$NN-variants it outperformed was the Generalised Mean Distance $K$NN, which finds the $K$ closest training samples from every class $m$, converts lists of these samples to local mean values, and then computes multiple mean distances to obtain distances between $\mathbf{x}$ and all $M$ classes~\cite{Uddin2022}. In~\cite{Minaev2018} the Euclidean distance was compared to 66 alternative distance measures, such as Minkowski and Manhattan distances, inside a $K$NN for indoor localization based on ion-mobility spectrometry fingerprints.

The choice of $K$ has been target of numerous studies. For fixed $K$ all neighbors of the test sample would converge towards $\mathbf{x}$ if $N$ would converge to infinity~\cite[p.\ 183]{Duda2001} and any value of $K$ would be acceptable. However, in the real world $N<< \infty$ and choosing a suitable $K$ is not straightforward. 

For $K \rightarrow \infty$ the $K$ nearest neighbors rule would be optimal\footnote{For a two-category classification problem the error rate converges to the lower bound, known as the Bayes rate if $K \rightarrow \infty$~\cite[p.\ 184.]{Duda2001}.}~\cite[p.\ 183]{Duda2001} as it eliminates any measurement noise, so a large $K$ is desirable. The drawback is that for large $K$ classes with large numbers of samples would be preferred over classes with small numbers of samples~\cite{Onyezewe2021}, which could weaken the classification accuracy. Therefore, a small $K$ would be desirable to ensure that all nearest neigbors are close to $\mathbf{x}$, but here class outliers might cause false labelling~\cite{Onyezewe2021}. Furthermore, having the same or at least similarly number of samples for each class is desirable, but still a compromise needs to be found. One option would be to calculate the error rates for a large range of feasible $K$ values and then choose the one with the minimal error rate.\footnote{What values are feasible depends, for example, on the total number of training samples $N$ and classes $M$.}

In general, an odd value is chosen for $K$ as it limits the large-sample two-class error rate from above~\cite[p.\ 183]{Duda2001} and helps to avoid ties in the majority vote in classification problems where the $K$ closest training samples are from two classes. This does not ensure, however, that the $K$NN always finds a label. Consider the case where $K=9$ and $M>3$. Then it would be possible that the nine nearest training samples consist of three times three samples from three different classes, which would result in a tie. One approach to avoid such ties is using weights $w_k$ ($k = \{1, 2, ..,K\}$) for the nearest neighbors such that $\sum_{k=1}^{K} w_k = 1$. A common approach is to use the inverses of distances between test sample and nearest neighbors and normalize them to sum up to one.

Besides finding optimal $K$ values that would be used for classification of any test sample independent on its location and the training samples in its neighborhood, attempts have been made to use flexible $K$ values. For example, Wettschereck and Dietterich~\cite{Wettschereck1993} proposed four $K$NN variants that determine optimal $K$ values for each training sample. Their first approach stores for each training sample a list of $K$ values that would correctly classify the training sample using leave-one-out cross-validation. For classifying a test sample $\mathbf{x}$ its $\tilde{K}$ nearest neighbors are searched. Next $K_\text{opt}$, which is the $K$ value that would correctly classify most of the $\tilde{K}$ neighbors, is determined. $K_\text{opt}$ is then used to determine the label of $\mathbf{x}$ based on majority vote as in the standard $K$NN. Thus, the $K_\text{opt}$ differs locally, and hence is called by the authors a locally adaptive $K$NN method. The three remaining $K$NN variants in~\cite{Wettschereck1993} are modifications of this adaptive $K$NN. All four variants are suitable for applications in which patterns differ considerably for different regions. For example, in~\cite{Martinez2021} $K$ values estimated to be optimal ranged from 1 to 10 for regions with low training sample density, while for regions with high intensity optimal $K$ values ranged from 30 to 50. Numerous adaptive $K$NN variants have been proposed since the publication of~\cite{Wettschereck1993}. One example is the locally adaptive $K$NN based on discrimination class (DC-LAKNN), which first determines the discrimination classes of the first and second majority class for various $K$ values. It then uses quantity and distribution information in the discrimination classes to find the optimal $K$ to be used in classification~\cite{Pan2020}.

In~\cite{Hassanat2014} authors proposed a $K$NN-variant based on an ensemble approach. It classifies the test sample $\sqrt{N}$ times with $K$NN classifiers for which $K = \{1, 2, .., \sqrt{N} \}$. The overall label is obtained by using the weighted sum of the results from the $\sqrt{N}$ $K$NNs.\footnote{In case $\sqrt{N} \notin \mathbb{N}$ then $\sqrt{N}$ rounded to the nearest integer towards $-\infty$ is used as upper value for $K$.} In~\cite{Wang2006} a $K$NN variant that demands the user to define a confidence level $p$ ($p \in [0,1]$) rather than $K$ is proposed. The algorithms adjusts $K$ such that the probability for the label of $\mathbf{x}$ to be correct is at least $p$. In~\cite{Cheng2014} sparse learning is used. The proposed $K$NN-variant accounts for correlation between samples and reconstructs $\mathbf{x}$ as a linear combination of training samples before determining the label by majority vote. The drawback of most methods that search for optimal $K$ values is that the optimization process is often time-consuming.

To avoid the need for optimization Fuzzy $K$NN-variants could be used, which provide probabilities of $\mathbf{x}$ belonging to any of the classes present amongst the $K$ nearest neighbors. Thus, it is less likely to get ties and similarly to the approach in~\cite{Wang2006} a confidence level on the overall label is available. However, even for Fuzzy $K$NN identifying a suitable $K$ is nontrivial and if training samples are uncertain or ambiguous then using a fixed $K$ is unreliable~\cite{Bian2022}. Therefore, \cite{Bian2022} proposes to derive an optimal $K$ for any test sample, increasing the computational demand considerably. In~\cite{Zhang2017} the standard $K$NN is extended by a training stage in which a decision tree is build for predicting an optimal $K$ for any test sample. In the classification stage first the optimal $K$ is derived from the decision tree for a test sample, before the standard $K$NN with $K$ set to the optimal value is used to obtain the label for the test sample. In~\cite{Montoliu2022} the N-Tuple Bandit Evolutionary Algorithm~\cite{Lucas2018} is used to obtain the optimal combination of $K$ and distance metric or a combination close to the optimum.

Various further $K$NN-variants have been proposed. The reader is referred to surveys (e.g.,~\cite{Uddin2022,Jiang2007}) that provide a thorough overview on the topic. The algorithm proposed in the next section does not require any optimization of $K$. Instead it works exactly like the standard $K$NN except for requiring a different number as input.


\section{Flexible K Nearest Neighbors}\label{sec:FlexKNN}

One limitation of the $K$NN-variants mentioned above is that they assume that a test sample belongs to one of the classes for which training data is available. However, this assumption does not hold in the real world all the time. Furthermore, many variants aim at optimizing the value of $K$ without considering the distances between nearest neighbors and the test sample for anything other than weighting the impact of the nearest neighbors on the final label for the test sample.

\begin{figure*}
\includegraphics[width=\textwidth,clip=true, trim=8cm 14cm 7cm 14cm]{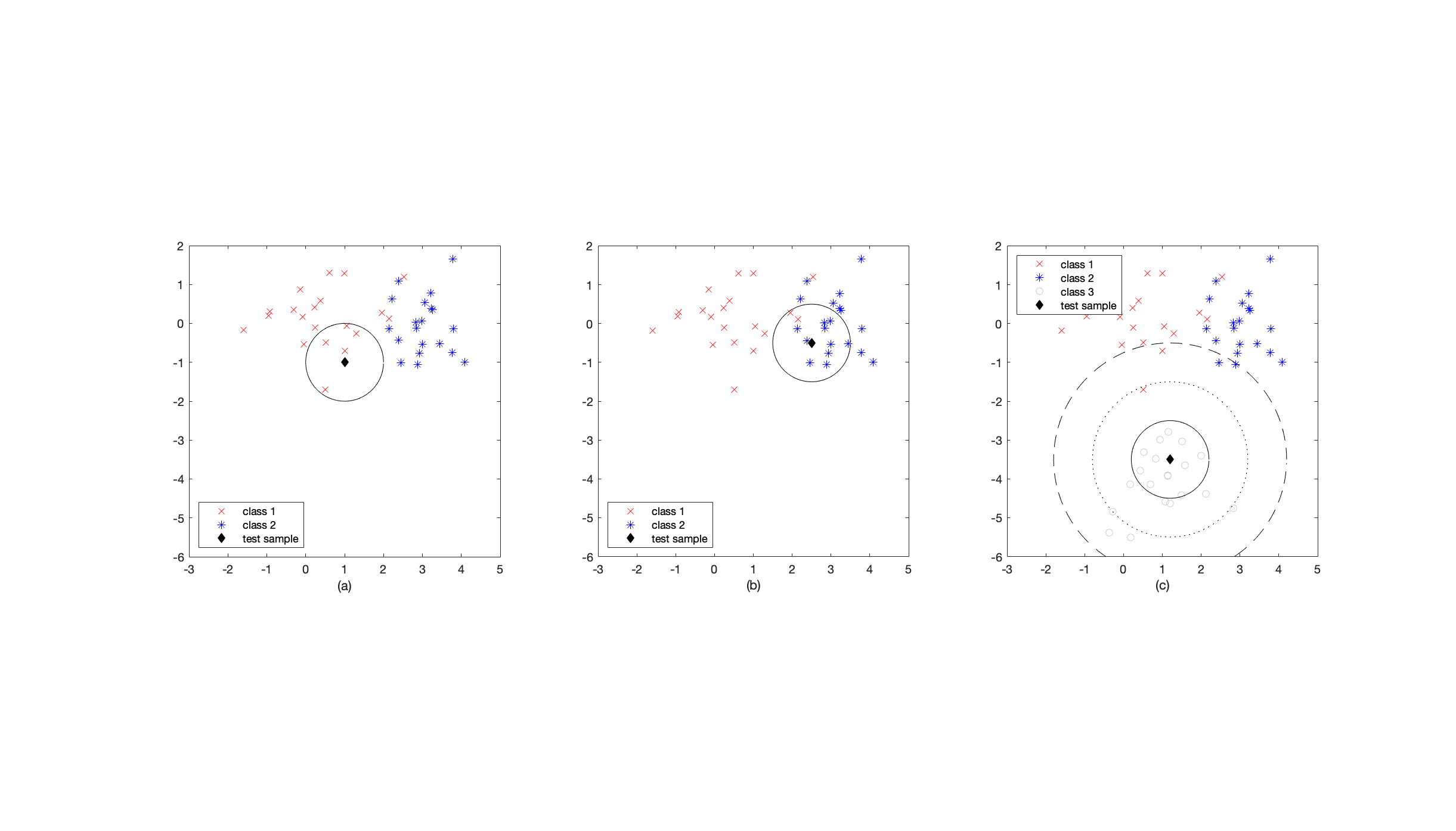}
\caption{
Illustration of the working principle of the Flexible $K$NN for two-dimensional data.Training samples from two classes are visualised by red crosses (class 1) and blue asterisks (class 2) in all three subfigures. The maximum allowed distance $d_\text{max}$ between test sample (black diamond symbol) is set to one and is visualized by the black circle around the test sample. The Flexible $K$NN classifies the test sample belonging to class 1 in (a) or class 2 in (b). For the $K$NN the yielded label depends on the choice of $K$. In subfigure (c) samples from a third class are visualized by grey circles. These samples are unknown to the classifier. The test sample clearly belongs to this unknown class. The Flexible $K$NN returns information that no label can be provided because no training sample lies within $d_\text{max}$. Contrary, the $K$NN would classify the test sample wrongly as belonging to either class 1 or 2 depending on the choice of $K$.}
\label{fig:FlexKNN}
\end{figure*}

Figure~\ref{fig:FlexKNN} illustrates why this approach is not appropriate all time. Here all samples are assumed, for simplicity, to be $n=2$. However, it is straightforward to extend the example to higher dimensional spaces. In Figure~\ref{fig:FlexKNN}(a) and (b) training samples from two classes are visualised by red crosses (class 1) and blue asterisks (class 2), and each class consists of 20 training samples. The test sample is visualized by the black diamond. The circle around it has a radius of one. The standard $K$NN would classify the test sample in Figure~\ref{fig:FlexKNN}(a) as belonging to class 1 for any $K=\{1,2, .., 5\}$. For the test sample in Figure~\ref{fig:FlexKNN}(b) $K$NN would classify it as belonging to class 2 for, at least, any $K=\{1,2, .., 12\}$ (there are twelve training samples inside the black circle, with ten of them belonging to class 2). Figure~\ref{fig:FlexKNN}(c) now contains also samples from class 3 illustrated by 20 grey circles. Clearly the test sample would now belong to class 3 (eleven training samples from class 3 are inside the circle with radius one). However, if these training samples were not available to the classifier then the standard $K$NN and the variants presented before would label the test sample as being a member of either class 1 or class 2, depending on the choice of K and whether distance-based weights are used.

In order to prevent such a misclassification, this paper introduces a $K$NN variant that does not use $K$ as a parameter chosen by either the user or an optimization algorithm. Instead the maximum distance $d_\text{max}$ between test sample and training samples is used as the only input parameter. Thus, the value of $K$ is here an output rather than input parameter as in the $K$NN. It varies for each test sample and depends on the locations of the test sample and the training samples in the $d$-dimensional space. Hence, the proposed $K$NN-variant is called Flexible $K$NN (Flex$K$NN).

Let us revisit the examples in Figure~\ref{fig:FlexKNN} and assume that $d_\text{max} = 1$ based on prior knowledge on the training data. In the example Figure~\ref{fig:FlexKNN}(a) five training samples from class 1 are inside the circle with radius $d_\text{max}$. Thus, $K=5$ and Flex$K$NN would classify the sample as being a member of class 1. In the example in Figure~\ref{fig:FlexKNN}(b) ten samples from class 2 and two from class 1 are inside the circle, hence $K=12$ and Flex$K$NN would classify the sample as being a member of class 2. For the example in Figure~\ref{fig:FlexKNN}(c) no training samples are inside the circle with radius $d_\text{max}$.\footnote{Grey dots illustrate training samples that are not available to the classifier.} Hence, Flex$K$NN would return information that the test sample is either an outlier or does not belong to either of the two classes for which training samples are available to the classifier. Only by increasing $d_\text{max}$ to 2 (dotted circle) one training sample from class 1 would be inside the maximum allowed distance. However, based on the distances of this red cross to other red crosses it might be an outlier itself. Further increasing $d_\text{max}$ to 3 (dashed circle) would result in two samples from each class 1 and class 2 inside the circle. In such a situation the normalized, inverse distance between test sample and training samples could be used as weights.

Algorithm~\ref{alg:FlexKNN} provides the pseudo-code for the Flexible $K$NN. The only difference compared to the standard $K$NN is that instead of $K$ maximum distance $d_\text{max}$ is required as input parameter. As mentioned before the maximum distance could be chosen randomly, but it is advisable to use prior knowledge. For example, one could calculate for each class the average distance over all training samples from class $l$ using leave-one-out cross-validation. This means, for each sample $x_i$ ($i=\{1,2, .., N_l\}$) in the class distances to samples $x_j$ ($j=\{1,2, .., N_l\}$, $j \not = i$) are calculated and the average distance $d_i$ for sample $x_i$ is stored. Finally, the average over all $d_i$ ($i=\{1,2, .., n\}$) is calculated to obtain $d_\text{l}$. Assuming that there are $M$ classes $d_\text{max} = \frac{1}{m} \sum_{l=1}^{M} d_\text{l}$. Alternative definitions for $d_\text{max}$, such as the median over all $d_\text{l}$, are also possible.

\begin{algorithm}
\begin{algorithmic}
\STATE \textbf{Input:} training samples $\mathbf{Y} = [\mathbf{y}_1, \ldots , \mathbf{y}_N] \in \mathbb{R}^{n \times N}$ from $M$ classes, test sample $\mathbf{x} = [ x_{1}, \ldots , x_{n} ] \in \mathbb{R}^n$, maximum distance between test and training samples $d_\text{max}$.
\STATE \textbf{Output:} label for test sample $\mathbf{x}$, number of training samples $K$ within distance $d_\text{max}$ of test sample $\mathbf{x}$.
\STATE \textit{Step 1:} Calculate Euclidean distances $d_i$ between $\mathbf{x}$ and $\mathbf{y}_i$ ($i = \{1, \ldots N\}$) using equation~(\ref{eq:EuclideanDistance}).
\STATE \textit{Step 2:} Search all training samples for which $d_i \leq d_\text{max}$. Training samples fulfilling the condition form the set of $K$ nearest neighbors.
\STATE \textit{Step 3:} Determine label of $\mathbf{x}$ based on the labels of the $K$ nearest neighbors using majority vote. If $K=0$, return information that no training samples are within $d_\text{max}$ from $\mathbf{x}$ and that no label can be provided.
\end{algorithmic}
\caption{Pseudo-code of Flexible $K$NN}
\label{alg:FlexKNN}
\end{algorithm}

Although Euclidean distance yields in general good performance, alternative distance measures could be employed in Algorithm~\ref{alg:FlexKNN} to potentially improve accuracy levels (for a thorough overview on distance measures see, e.g.~\cite{DezaBook}). For example, in~\cite{Minaev2018} 66 alternative distance measures were compared to the Euclidean distance for indoor localization based on IMS fingerprints. Ruzicka, Canberra, and Vicis Symmetric $\chi^2$ were Pareto optimal in the tests and achieved higher localization accuracy than the Euclidean distance while requiring less computation time. However, other metrics can require additional parameters to be defined by the user.


\section{Application for IMS-based localization}\label{sec:Appl}

\subsection{Test with full training data}

This section compares the performances of the proposed Flex$K$NN and standard $K$NN for positioning based on IMS fingerprints. For evaluation the dataset from~\cite{WPNC2017} is used, as the $K$NN performed in that paper poorly for part of the test samples due to training and test samples being too dissimilar. The Flex$K$NN is believed to mitigate this problem.

Ion-mobility spectrometry is a technique for measuring volatile organic compounds (VOCs). To the authors knowledge, the possibility to localize based on VOC fingerprints was studied for the first time in~\cite{WPNC2017}. The dataset contained 8,736 IMS fingerprints from seven different rooms on the campus of Tampere University of Technology, Finland. For each room data was collected once during weekend when the university buildings were (almost) empty and once during the week when staff and students were present. This was done to investigate the temporal stability of IMS fingerprints. It is known that especially humidity and temperature, but also air currents and barometric pressure influence the mobility of molecules~\cite[p.\ 250 ff.]{Eiceman2014} and thus have an impact on the IMS readings.

The data from~\cite{WPNC2017} were collected using a ChemPro100i from Environics Oy (Mikkeli, Finland), which yielded IMS fingerprints of dimension $n=14$. For each room approximately 600 fingerprints were collected for both empty (ie., during weekend) and crowded (i.e., on a weekday) conditions.

In~\cite{WPNC2017} the $K$NN performed well (classification accuracies close to 100\%) for $K = \{1, 3, 5, 7\}$ when training and test samples were collected on the same day but performed poorly (classification accuracies between 28.21\% and 37.38\%) when training and test data were collected on different days, in different conditions. The analysis confirmed that IMS fingerprints depend strongly on the environmental conditions and normalizing{\footnote{Data was normalized in~\cite{WPNC2017} by subtracting the mean and dividing by the standard deviation of all measurements for each of the 14 dimensions.}} the data was insufficient to mitigate the impact.

\subsection{Test with full training data}

\begin{figure}
\centerline{\includegraphics[width=.48\textwidth,clip=true, trim=0cm 1cm 0cm 1.8cm]{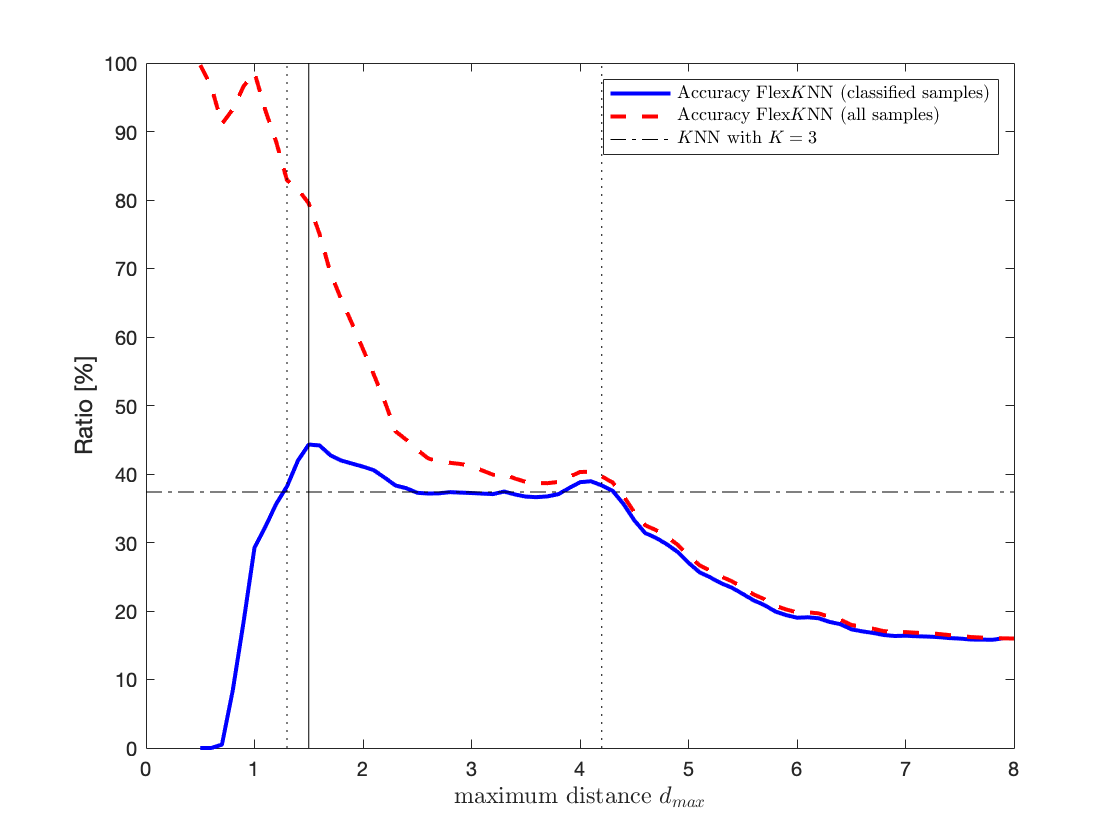}}
\caption{Classification accuracy of Flex$K$NN for varying $d_\text{max}$. The blue, solid line shows ratio of accurately classified test samples for which Flex$K$NN returned a label. The red, dashed line shows the ratio of test samples that were either correctly classified or for which Flex$K$NN did not yield a label because no training sample was within $d_\text{max}$. For comparison the classification accuracy of the $K$NN with $K=3$ is shown (black, dashed-dotted line).}
\label{fig:AccFlexKNNvs3NN}
\end{figure}

For evaluating Flex$K$NN and standard $K$NN in this section normalized data from crowded (4,375 samples) and empty conditions (4,361 samples) were used for training and testing respectively. Distances between samples were measured by the Euclidean distance~(\ref{eq:EuclideanDistance}). The $K$NN with $K=3$ classified 37.38\% of the training samples correctly. For the Flex$K$NN the classification performance was checked for $d_\text{max} = \{0.1, 0.2, .., 8\}$ and the accuracies are shown in Figure~\ref{fig:AccFlexKNNvs3NN}. The blue line illustrates the ratio of test samples for which the Flex$K$NN yielded a correct label. Between $d_\text{max} = 1.3$ and $d_\text{max} = 4.2$ (vertical dotted lines) the accuracy level is higher or approximately the same than that of the $K$NN (dash-dotted black line), with the highest accuracy at $d_\text{max}=1.5$ (vertical solid line). However, the more important line is the red, dashed line in Figure~\ref{fig:AccFlexKNNvs3NN}. It illustrates the ratio of test sample for which the Flex$K$NN yielded either the correct label or returned information that no training samples within $d_\text{max}$ were found and no label could be provided. In the latter case the $K$NN yielded a label, but its trustworthiness was low and this resulted, in general, in a misclassification.

As $d_\text{max}$ increases the red and blue lines converge. This indicates that the ratio of test samples that cannot be classified due to missing training samples within $d_\text{max}$ decreases. For $d_\text{max}=8$ all test samples were classified, but the classification accuracy of the Flex$K$NN already dropped at $d_\text{max}=2.5$ below that of the $K$NN. This can be, partly, explained by the fact that the number of training samples within $d_\text{max}$ steadily increase as $d_\text{max}$ increases and that it is considerably higher than the $K$ value usually used inside the $K$NN. For example, at $d_\text{max} = 1.5$ the average $K$ over all 4,361 test samples was already 181.24 and at $d_\text{max} = 4.2$ it was 2,176.83. Considering that for each room there were only roughly 600 training samples it is not surprising that the classification accuracy decreases once $d_\text{max}>4.2$. For $d_\text{max} = 4.2$ at most $\approx$28\% of the training samples inside $d_\text{max}$ could be from the correct class, making a misclassification likely. To conclude, for the presented dataset choosing $d_\text{max} \approx 1.5$ provides the best compromise between high accuracy of provided labels and low number of non-classified test samples.

\begin{figure}
\centerline{\includegraphics[width=.48\textwidth,clip=true, trim=0cm 1cm 0cm 1.8cm]{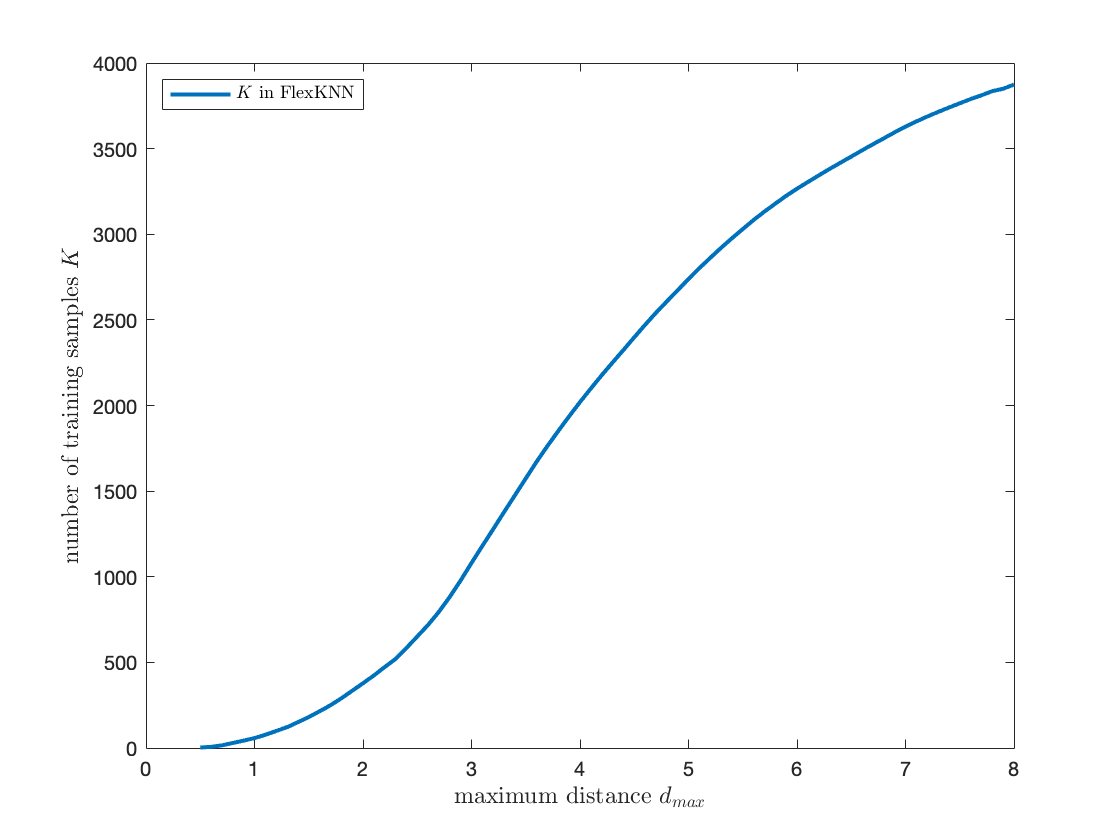}}
\caption{Number of training samples $K$ inside $d_\text{max}$ for Flex$K$NN. Label of test sample is derived from labels of these $K$ training samples.}
\label{fig:KinFlexKNN}
\end{figure}


\subsection{Test with missing room data}

For the data from~\cite{WPNC2017} it was noted that IMS measurements differed most noticeably for training and test samples collected in rooms 6 and 7. Therefore, in this section only training samples from rooms 1 to 5 were used for determining the labels of the same 4,361 test samples as in Section~\ref{sec:Appl}-A, of which 1,234 were collected in rooms 6 and 7. For the $K$NN $K=3$ and for the Flex$K$NN $d_\text{max}=1.5$.

The classification accuracy of the $K$NN was 71.70\%, which is a clear improvement over the result in Section~\ref{sec:Appl}-A and supports the hypothesis that larger differences in the samples from rooms 6 and 7 collected for training and testing caused a large portion of the misclassifications. The Flex$K$NN yielded for 3,648 test samples a label, of which 76.75\% were correct; for the remaining 713 test samples no training sample was within $d_\text{max}$ (i.e., $K$=0) and hence the Flex$K$NN did not return a label. The overall accuracy, which accounts for test samples being correctly classified as belonging to rooms 1 to 5 or having no label due to $K$=0, was 80.55\%.

Test samples without label stem mostly from rooms 6 and 7. However, 42.22\% of the samples from these two rooms were also misclassified by the Flex$K$NN\footnote{The $K$NN misclassified all samples from rooms 6 and 7 due to missing training data from these two rooms.}, which indicates that either a smaller $d_\text{max}$ should be used or that some IMS fingerprints between rooms 6 and/or 7 and one of the remaining five rooms were too similar to be distinguished reliably. It also shows that choosing a suitable $d_\text{max}$ is non trivial. A thorough analysis of the training data investigating the density of training samples, the closeness of training samples from the same class as well as the closeness of training samples from different classes could help find the $d_\text{max}$ that yields the highest accuracy.


\section{Concluding remarks}\label{sec:Concl}

This paper introduced a modification of the widely used $K$ Nearest Neighbors classifier for which the maximum allowed distance between the sample to be classified (test sample) and training samples is used as input parameter. Hence, $K$ is flexible and can differ considerably between different training samples. Consequently, the algorithm was named Flexible $K$NN (abbreviated as Flex$K$NN). The reasoning behind the Flex$K$NN is that the standard $K$NN and its variants will always yield a label for a test sample even if the $K$ closest training samples are far away from the test sample. This might occur, for example, if the test sample is from a class for which no training data is available. In such scenario existing $K$NN-variants will yield a wrong label while the Flex$K$NN will provide information that no label could be determined due to all training samples being too dissimilar and that it is reasonable to assume that the test sample belongs to a yet unknown class.

In Section~\ref{sec:Appl}, the Flex$K$NN was compared to the standard $K$NN for localization based on ion-mobility spectrometry fingerprints. The dataset from~\cite{WPNC2017} was chosen because it highlighted the limitation of the $K$NN, in situations where training samples were very dissimilar to the test sample, and the capability of the Flex$K$NN to solve or at least mitigate this limitation.The test showed that the Flex$K$NN can outperform the $K$NN for reasonable choices of $d_\text{max}$, even when training data are available for all classes observed in the test data.

The maximum allowed distance $d_\text{max}$ between the test sample and training samples can be determined, for example, from the training data using leave-one-out cross-validation or by simply testing the Flex$K$NN's performance for various values of $d_\text{max}$ to find the one yielding the highest accuracy. In future work also techniques for systematically or dynamically determining the optimal $d_\text{max}$ will be studied. Alternatively, other prior information on the data could be used. An advantage of the Flex$K$NN is that the number of training samples $K$ inside $d_\text{max}$ provides information how well the test sample fits into the existing clusters of training data. Large $K$ and/or $K$ closest training samples mostly from one class indicate a good fit and high confidence in the yielded label; low $K$ and/or $K$ closest training samples from various classes indicate a poor fit and low confidence in the yielded label. A reasonable requirement, to avoid the latter case, would be to demand that $K$ is above a certain threshold to avoid misclassification based on some outlier data.

Several methods that have been proposed for improving existing $K$NN-variants have been discussed in Section~\ref{sec:RelWorks}. Part of their ideas could also be used within the Flex$K$NN. For uneven class distribution it would be advisable to calculate the ratio of training samples from a certain class inside $d_\text{max}$~\cite{Onyezewe2021} and either choose the label based on the class with the highest ratio or use normalized ratios of all classes to assign the test samples fuzzy memberships for all classes. In order to avoid ties the concept of the weighted $K$NN could be used, meaning that each of the samples inside the maximum allowed distance  is given a weight that is inverse proportional to its distance to the test sample.

The usefulness of all these potential modifications to the Flex$K$NN will be tested in the future with various datasets. Also, a thorough comparison with other $K$NN-variants will be carried out. This comparison will provide an insight into the pros and cons of each $K$NN-variant, their computational efficiency, and in which scenarios to use which variant.


\section*{Acknowledgment}

The author thanks Anton Rauhameri for proofreading and commenting the manuscript as well as Dr. Simo Ali-Löytty for discussing $K$ Nearest Neighbors and the proposed approach.

\vspace{12pt}
\end{document}